\documentclass[10pt,twocolumn,letterpaper]{article}

\usepackage{iccv}              % To produce the CAMERA-READY version
\usepackage{multirow}
\usepackage{algorithm2e}
\usepackage{dsfont}
\usepackage{amsthm}
\usepackage{pifont}

\newtheorem{remark}{Remark}
\newtheorem{theorem}{Theorem}[section]

\definecolor{iccvblue}{rgb}{0.21,0.49,0.74}
\usepackage[pagebackref,breaklinks,colorlinks,allcolors=iccvblue]{hyperref}

\def\paperID{9881} % *** Enter the Paper ID here
\def\confName{ICCV}
\def\confYear{2025}

\title{More Reliable Pseudo-labels, Better Performance: A Generalized Approach to Single Positive Multi-label Learning}

\author{Luong Tran\\
FPT Software AI Center\\
{\tt\small luongtk@fpt.com}
\and
Thieu Vo\\
National University of Singapore\\
{\tt\small thieuvo@nus.edu.sg}
\and
Anh Nguyen\\
University of Liverpool\\
{\tt\small anh.nguyen@liverpool.ac.uk}
\and
Sang Dinh\\
Hanoi University of Science and Technology\\
{\tt\small sangdv@soict.hust.edu.vn}
\and
Van Nguyen\\
FPT Software AI Center\\
{\tt\small vannth19@fpt.com}
}

\begin{document}
\maketitle
% \begin{abstract}
% Multi-label learning is a challenging computer vision task that associates each image with multiple categories. However, fully annotating large datasets with multiple labels per image is often impractical due to the high costs and effort required. This challenge motivates the exploration of multi-label learning with partially annotated data, with an extreme case being Single Positive Multi-Label Learning (SPML), a scenario where each instance is annotated with only one positive label and the other labels are unannotated. Traditional SPML approaches, which treat missing labels as either unknown or negative, often lead to inaccuracies or false negatives. To address these challenges, we leverage Vision-Language Models (VLMs), known for their excellent zero-shot classification capabilities, in order to improve supervision in SPML. We propose the Adaptive and Efficient Vision-Language Pseudo-Labeling (AEVLP) framework, which comprises a novel Dynamic Augmented Multi-focus Pseudo-labeling (DAMP) technique and Generalized Pseudo-Label Robust Loss (GPR Loss). Specifically, multiple dynamic focuses of each training image are presented to the VLM to produce more reliable pseudo-labels, and the GPR Loss facilitates effective learning from these pseudo-labels and mitigates the noise they may introduce. Extensive experiments demonstrate that our framework achieves significant advancements in multi-label classification, setting state-of-the-art results on four benchmark datasets.
% \end{abstract}

\begin{abstract} Multi-label learning is a challenging computer vision task that requires assigning multiple categories to each image. However, fully annotating large-scale datasets is often impractical due to high costs and effort, motivating the study of learning from partially annotated data. In the extreme case of Single Positive Multi-Label Learning (SPML), each image is provided with only one positive label, while all other labels remain unannotated. Traditional SPML methods that treat missing labels as unknown or negative tend to yield inaccuracies and false negatives, and integrating various pseudo-labeling strategies can introduce additional noise. To address these challenges, we propose the Generalized Pseudo-Label Robust Loss (GPR Loss), a novel loss function that effectively learns from diverse pseudo-labels while mitigating noise. Complementing this, we introduce a simple yet effective Dynamic Augmented Multi-focus Pseudo-labeling (DAMP) technique. Together, these contributions form the Adaptive and Efficient Vision-Language Pseudo-Labeling (AEVLP) framework. Extensive experiments on four benchmark datasets demonstrate that our framework significantly advances multi-label classification, achieving state-of-the-art results. \end{abstract}    

\section{Introduction}
\label{sec:intro}
Most visual classification studies focus on multi-class settings which aim to identify the most suitable label for a given input image from a set of possible options. However, in real-world scenarios, an image often contains multiple objects or attributes that can be considered as their labels. Therefore, multi-label learning \cite{liu2021emerging, zhang2013review} has emerged as a solution to develop predictive models capable of assigning multiple labels to an unseen image. In early stages, multi-label learning required annotation of all relevant classes for each training instance; however, such an exhaustive annotation scheme is often costly and impractical \cite{deng2014scalable}.

Due to the cost and impracticality of exhaustive annotation \cite{deng2014scalable}, researchers have explored multi-label learning with missing labels \cite{6977055, yu2014large, wu2015ml, zhang2021simple}. In its extreme setting, known as Single-Positive Multi-Label Learning (SPML) \cite{cole2021multi}, each image is annotated with only one confirmed positive label, leaving the remaining labels unknown. This approach significantly reduces labeling cost and demonstrates that multi-label classifiers can be learned with minimal supervision.

A common strategy in SPML is to treat missing labels as negative (the Assume Negative method \cite{cole2021multi}), and many existing methods \cite{chen2024boosting, xu2022one, zhou2022acknowledging, liu2023revisiting} further rely on the model's internal knowledge to generate pseudo-labels for fully supervised learning. However, this approach often produces inherently noisy pseudo-labels, leading to repeated errors during training, an issue that prior studies have not effectively addressed. To tackle this challenge, we propose the Generalized Pseudo-label Robust Loss (GPR Loss), a novel loss function designed to learn effectively from pseudo-labels generated by an external knowledge-based method while robustly mitigating the noise associated with uncertain labels.

Recently, the emergence of Vision-Language Models (VLMs) has enhanced performance in various recognition tasks \cite{zhang2024vision}, including multi-label learning. One approach involves fine-tuning these models with additional network layers \cite{ding2023exploring}, while another leverages model outputs with pseudo-labels in an unsupervised, annotation-free manner \cite{abdelfattah2023cdul}. Recent work \cite{xing2024vision} generates a fixed pseudo-label vector for each image using a pretrained large CLIP model. However, relying on a single fixed pseudo-label vector prevents the model from re-evaluating and correctly associating an image with categories that were initially misclassified as false negatives during training.

Based on these observations, to verify the strength of GPR Loss, we introduce a simple yet effective dynamic pseudo-labeling process that encourages CLIP to attend not only to the global image but also to various local areas, which are randomly selected and augmented during training. This simple pseudo-labeling approach allows the pseudo-label vector for each image to vary from epoch to epoch, reducing the chance of missing correct associations.

In summary, the main contributions of this work are twofold. \textit{First}, we propose the Generalized Pseudo-label Robust Loss (GPR Loss) to effectively learn from pseudo-labels generated by an external knowledge-based method while mitigating the noise from uncertain labels. \textit{Second}, we introduce the Dynamic Augmented Multi-focus Pseudo-labeling (DAMP) technique to generate reliable pseudo-labels in a stable manner, which, together with GPR Loss, forms the AEVLP framework. Through extensive experiments, to the best of our knowledge, we demonstrate that this framework achieves state-of-the-art (SOTA) results for the SPML problem compared to existing methods based on loss design and pseudo-labeling.
\section{Related Works}
\label{sec:related_works} 

%Multi-label learning involves developing models to predict multiple labels related to unannotated images. Typically, deep neural networks are trained with multiple outputs, each predicting the presence or absence of specific classes. In scenarios where all labels are available for each image, standard binary cross-entropy are commonly applied \cite{cole2021multi}. However, obtaining fully and accurately labeled training instances can be prohibitively expensive and impractical. This limitation has inspired researchers working on Multi-label Learning with Missing Labels \cite{6977055}, which addresses the challenge of learning with incomplete label information and is considered a form of weakly supervised learning \cite{kim2022large}.

%This study focuses on Single Positive Multi-label Learning (SPML), an extreme variant of Multi-label Learning with Missing Labels where each image is known to have only one verified positive label, with all other labels considered unknown \cite{cole2021multi}. SPML presents a practical yet highly challenging scenario compared to fully labeled settings. 
% Existing methods for general multi-label learning with missing labels, such as learning positive label correlations \cite{huynh2020interactive}, creating label matrices \cite{feng2020regularized}], or inferring missing labels \cite{durand2019learning}, face significant obstacles when dealing with a single positive label \cite{cole2021multi}.

In SPML, two strategies are employed to handle missing labels. The first strategy treats missing labels as unknown variables to be predicted \cite{rastogi2021multi}. The second strategy assumes that all missing labels are negative, transforming SPML into a fully supervised multi-label learning problem \cite{cole2021multi, xu2022one, zhou2022acknowledging, kim2023bridging, kim2022large}. The Assume Negative assumption remains popular in multi-label learning with missing labels including SPML \cite{liu2023revisiting} and often serves as a baseline in benchmark experiments.

However, this naive assumption oversimplifies the problem and can lead to numerous false negative labels \cite{zhang2023learning}. To address this issue, 
\citeauthor{cole2021multi} \cite{cole2021multi} propose the online estimation of labels during training. Furthermore, \citeauthor{kim2022large} \cite{kim2022large} show that reducing large losses, which are potentially introduced by false negatives, can significantly improve performance. Meanwhile, the authors in \cite{kim2023bridging} modify the final stage of the CNN architecture to increase the attribution scores of output logits. In \cite{zhou2022acknowledging}, entropy maximization (EM) and asymmetric pseudo-labeling (APL) are combined to improve robustness. Additionally, \citeauthor{chen2024boosting} \cite{chen2024boosting} adapt class- and instance-wise loss concepts to SPML within a comprehensive framework. In \cite{xu2022one, liu2023revisiting}, the approach introduces a label enhancement process, enhancing model performance with minimal positive labels per instance. While our study continues to rely on the Assume Negative assumption and addresses the resulting false negatives, we also focus on learning from pseudo-labels and mitigating the noise they may introduce.

%While handling false negative labels can be a powerful method, further improvements in previous works become increasingly challenging, as they approach the upper bound of performance under minimal supervision.

To address the challenges caused by incomplete labeling in multi-label learning, the ability of the CLIP model \cite{radford2021learning} for zero-shot classification has emerged as an effective solution. Recent work by \cite{xing2024vision} forces the model to learn from pseudo-labels generated by CLIP. Another approach \cite{ding2023exploring} finetunes CLIP model with additional layers of a Graph Convolutional Network (GCN) \cite{kipf2017semisupervised}. Additionally, in \cite{abdelfattah2023cdul}, the approach is to generate soft pseudo-labels based on global-local image-text similarity aggregation. Our approach also leverages CLIP for pseudo-labeling; however, unlike previous methods, it employs dynamic pseudo-labeling throughout the training process, reassigning each image a \textit{new} set of pseudo-labels at every epoch. This helps mitigate pseudo-labeling errors while expanding the range of correct pseudo-labels. 

% In addition, unlike \cite{ding2023exploring}, where CLIP is a part of the model architecture in both training and inference, and the entire architecture is finetuned end-to-end, our framework uses a pretrained CLIP (or another similar model) solely as a means to generate pseudo-labels to improve supervision of a much simpler multilabel-classification backbone network. During inference, only this backbone network is present. Our framework not only simplifies inference but also training, allowing flexibility of paring different pseudo-labeling methods and backbone networks by disentangling the pseudo-labeling from the backbone image classification.

 % Additionally, prompt learning is employed to fine-tune CLIP in \cite{sun2022dualcoop, hu2023dualcoop++}.

% CLIP uses a large-scale dataset of images and associated textual descriptions to learn comprehensive visual and semantic representations. This model can generalize to novel, unseen classes without additional training by utilizing textual descriptions of the new classes. This capability is particularly beneficial for multi-label learning with missing labels, as it allows for the generation of pseudo labels to fill gaps where annotations are incomplete.%The versatility and robustness of CLIP in managing diverse and unseen categories make it a powerful tool for advancing traditional multi-label learning methodologies.

% Although this method can be a promising solution, it still has a problem: we cannot utilize advanced image classifier backbones if there is no corresponding CLIP model available.
\section{Problem Statement} 
% In this section, we introduce the AEVLP framework, a novel approach for SPML, as illustrated in \cref{fig:proposal}. AEVLP trains a model with pseudo-labels that change across epochs using a robust loss. During the training phase, the CLIP model and GCN remain frozen to mitigate computational costs and time. In the inference phase, only the main backbone network is utilized for predictions.

% To augment the zero-shot capabilities of the vision-language model, AEVLP incorporates the Augmentor, which systematically generates diverse augmented views of input images, and also integrates Graph Convolution Network (GCN) to introduce controlled noise into the embedding features encoded by the vision-language Text Encoder. 
% The loss function employed in AEVLP ensures that the model effectively learns from the pseudo-labels. 

\begin{figure*}[htbp]
    \centering        
    \includegraphics[width=\textwidth]{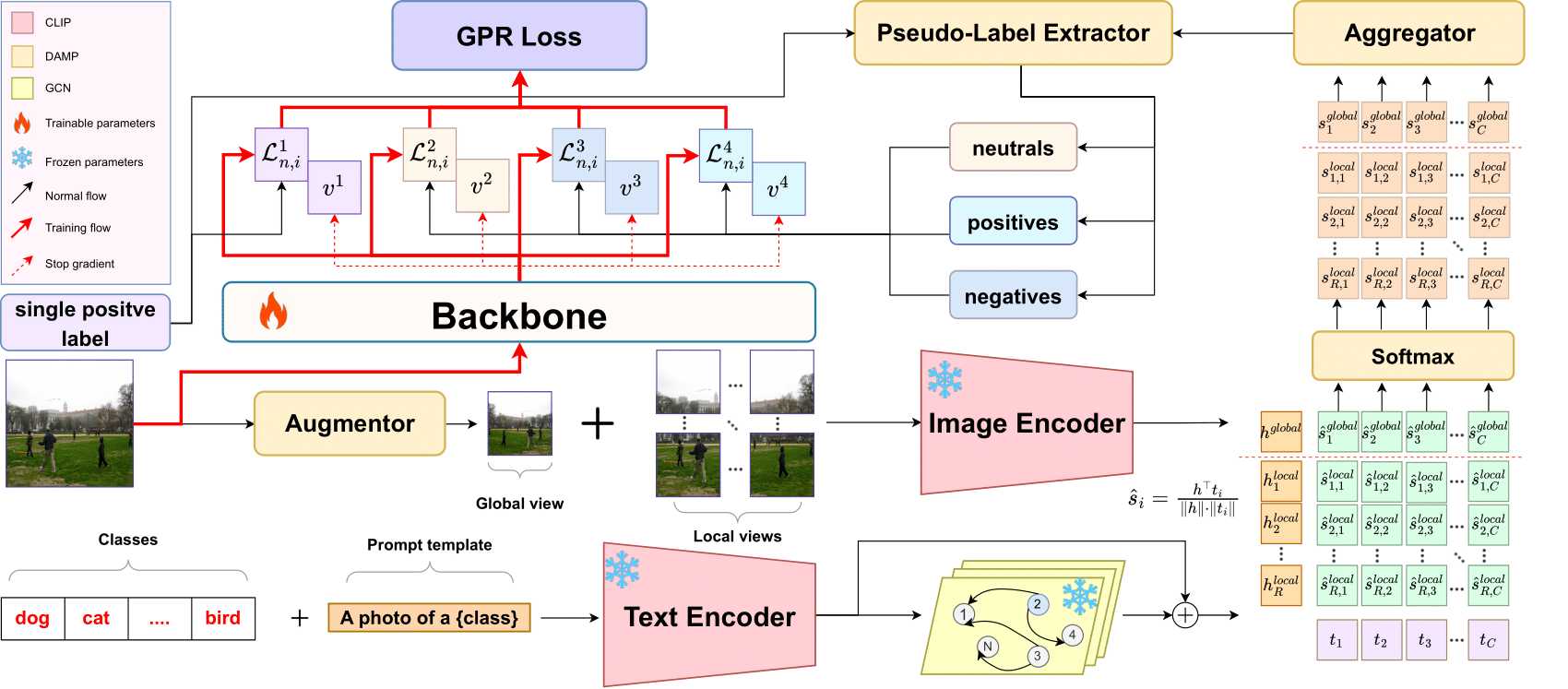}      \captionsetup{justification=centering, width=\textwidth}
    \caption[AEVLP Framework.]{An overview of the proposed method - Adaptive and Efficient Vision-Language Pseudo-Labeling Framework (AEVLP)}
    \label{fig:proposal}
\end{figure*} 

% \subsection{Problem Statement}  
\label{sec:problem_statement}
In the original multi-label learning problem, given a dataset \(\mathcal{D} = \{(x_n, y_n)\}_{n=1}^{N}\), where \(x_n \in \mathcal{X}\) (input space) and \(y_n \in \mathcal{Y} = \{0, 1\}^{C}\) (label space), and a validation dataset \(\mathcal{D}^{\text{val}} = \{(x^{\text{val}}_n, y^{\text{val}}_n)\}_{n=1}^{N'}\), the goal is to learn a mapping \(f: \mathcal{X} \rightarrow \mathcal{Y}\) that predicts multiple labels for each \(x_n\). We denote \( p_n = f(x_n) = \sigma(s_n) \), where \( s_n \) represents the model's logits and \( \sigma \) is the sigmoid activation function. The loss function is defined as follows:
\begin{equation*}
\label{eq:multi_bce}
\centering
\mathcal{L} = -\frac{1}{NC} \sum_{n=1}^N \sum_{i=1}^C y_{n,i}\log p_{n,i} + (1-y_{n,i})\log(1 - p_{n,i})
\end{equation*}

% The objective then becomes:

% \begin{equation}
% \label{eq:an}
% \centering
% \mathcal{L} = \frac{1}{NC} \sum_{n=1}^N \sum_{i=1}^C \hat{y}_{n,i}\log p_{n,i} + (1-\hat{y}_{n,i})\log(1 - p_{n,i})
% \end{equation}

To address the SPML problem, we apply the Assume Negative assumption \cite{cole2021multi}, which treats unobserved labels as negative, transforming the problem into the original multi-label learning setup. The original dataset \(\mathcal{D}\) is replaced with a pseudo-multi-label dataset \(\hat{\mathcal{D}} = \{(x_n, \hat{y}_n)\}_{n=1}^{N}\), where \(\hat{y}_n \in \hat{\mathcal{Y}} = \{0, 1\}^C\), ensuring \(\sum_{i=1}^{C} \hat{y}_{n,i} = 1\). Each \(\hat{y}_{n,i} = 1\) if \(x_n\) is relevant to class \(i\) (positive label) and 0 if the label is unknown or unannotated.

In SPML, based on pseudo-labeling, an instance can be represented as a triplet \((x_n, \hat{y}_n, l_n)\), where \(x_n\) is the input, \(\hat{y}_n\) is the label vector based on the Assume Negative assumption, and \(l_n \in \{-1, 0, 1\}^C\) represents pseudo-labels generated by an arbitrary method \(\mathcal{M}\) to approximate additional labels. For class \(i\), \(l_{n,i} = 1\) indicates a positive pseudo-label, \(l_{n,i} = -1\) a negative pseudo-label, and \(l_{n,i} = 0\) means the label is undefined by \(\mathcal{M}\).

\section{Generalized Pseudo-Label Robust Loss}

\subsection{Background}
\label{sec:gr_loss}
To address the noise introduced in the Assume Negative assumption, the authors in \cite{chen2024boosting} propose the Generalized Robust Loss (GR Loss) as follows: 
\begin{equation}
\label{eq:gr_loss}
\centering
    \mathcal{L}^{GR} = \frac{1}{NC} \sum_{n=1}^N \sum_{i=1}^C v^{old}(p_{n,i}; \alpha)\cdot \mathcal{L}_{n,i}^{old}
\end{equation} 
\noindent
where the probability of class \(i\) and instance \(n\) is denoted as \(p_{n,i}\); \(v^{old}(p_{n,i}; \alpha)\) and $\mathcal{L}_{n,i}^{old}$ are the class-and-instance-specific weight and loss, respectively. Let \(\mathds{1}_{[\cdot]}\) is the indicator function, the weight term \(v^{old}(p_{n,i};\alpha)\) is expressed as: 
\begin{equation*}
    v^{old}(p_{n,i}; \alpha) = \mathds{1}_{[\hat{y}_{n,i} = 1]}v^{1}(p_{n,i}; \alpha) + \mathds{1}_{[\hat{y}_{n,i} = 0]}v^{2}(p_{n,i}; \alpha), 
\end{equation*} 
where \(v^{1}(p; \alpha) = 1\) and \(v^{2}(p; \alpha) = \exp\left(-\frac{(p - \mu)^2}{2\sigma^2}\right) \). Here, \(\alpha = [\sigma, \mu]\) is updated linearly over training epochs, following \cite{chen2024boosting}, as detailed in the supplementary materials. The loss term \(\mathcal{L}^{old}_{n,i}\) is defined as: 
\begin{equation*}
    \mathcal{L}^{old}_{n,i} = \mathds{1}_{[\hat{y}_{n,i} = 1]} \mathcal{L}^{1}_{n,i} + \mathds{1}_{[\hat{y}_{n,i} = 0]} \mathcal{L}^{2}_{n,i},
\end{equation*}
where \(\mathcal{L}^1_{n,i} = -\log p_{n,i}\) and the term \(\mathcal{L}^2_{n,i}\) is computed as:
\[
    \mathcal{L}^2_{n,i} = (1 - \hat{k}(p_{n,i}; \beta)) \frac{1 - (1 - p_{n,i})^{q_1}}{q_1} + \hat{k}(p_{n,i}; \beta) \frac{1 - p_{n,i}^{q_2}}{q_2}.
\]
% \[
% \begin{aligned}
%     \mathcal{L}^1_{n,i} &= -\log p_{n,i}, \\
%     \mathcal{L}^2_{n,i} &= (1 - \hat{k}(p_{n,i}; \beta)) \frac{1 - (1 - p_{n,i})^{q_1}}{q_1} \\
%     &\quad + \hat{k}(p_{n,i}; \beta) \frac{1 - p_{n,i}^{q_2}}{q_2}.
% \end{aligned}
% \]
Here, \(\hat{k}(p_{n,i}; \beta)\) is the estimate of the label being a false negative during training with a parameter \(\beta\), while \(q_1\) and \(q_2\), treated as hyperparameters, balance the trade-off between MAE and BCE loss, as described in \cite{chen2024boosting} and illustrated in the supplementary materials. Note that both \(\hat{k}(p_{n,i}; \beta)\) and \(v(p_{n,i}; \alpha)\) are implemented with gradients stopped with respect to \(p_{n,i}\).

\subsection{Proposed Method}
\label{sec:gprloss}
The above method, GR Loss in \cref{sec:gr_loss}, only works with the naive Assume Negative assumption and cannot handle external pseudo-labels generated by a reliable method \(\mathcal{M}\). Inspired by \cite{xing2024vision,chen2024boosting}, we propose the Generalized Pseudo-Label Robust Loss (GPR Loss), as follows:
\begin{equation}
\label{eq:gpr_loss}
\centering
\mathcal{L}^{GPR} = \frac{1}{NC} \sum_{n=1}^N \sum_{i=1}^C v^{new}(p_{n,i}; \alpha) \cdot \mathcal{L}^{new}_{n,i} +\eta \mathcal{R}.
\end{equation}

We explain the terms in  \cref{eq:gpr_loss} below. The loss term \(\mathcal{L}^{new}_{n,i}\) and the weight term \(v^{new}(p_{n,i};\alpha)\) are expressed as follows:
\begin{equation*}
\begin{aligned}
\mathcal{L}^{new}_{n,i} = & \quad \mathds{1}_{[\hat{y}_{n,i} = 1]} \mathcal{L}^1_{n,i} + \mathds{1}_{[\hat{y}_{n,i} = 0 \land l_{n,i} = 0]} \mathcal{L}^2_{n,i} \\
& + \mathds{1}_{[\hat{y}_{n,i} = 0 \land l_{n,i} = -1]} \mathcal{L}^3_{n,i} + \mathds{1}_{[\hat{y}_{n,i} = 0 \land l_{n,i} = 1]} \mathcal{L}^4_{n,i},
\end{aligned}
\end{equation*}
\begin{equation*}
\begin{aligned}
v^{new}(p_{n,i}; \alpha) = & \quad \mathds{1}_{[\hat{y}_{n,i} = 1]} v^1(p_{n,i}; \alpha)\\
&+ \mathds{1}_{[\hat{y}_{n,i} = 0 \land l_{n,i} = 0]} v^2(p_{n,i}; \alpha) \\
& + \mathds{1}_{[\hat{y}_{n,i} = 0 \land l_{n,i} = -1]} v^3(p_{n,i}; \alpha) \\
&+ \mathds{1}_{[\hat{y}_{n,i} = 0 \land l_{n,i} = 1]} v^4(p_{n,i}; \alpha).
\end{aligned}
\end{equation*}

% \noindent where the new terms \(\mathcal{L}^{3}_{n,i}\) and \(\mathcal{L}^{4}_{n,i}\) handle negative and positive pseudo-labels, respectively. The functions \(v^3\) and \(v^4\) serve as weight functions that adjust the contribution of these losses based on the model’s prediction confidence.

The new loss terms \(\mathcal{L}^{3}_{n,i}\) and \(\mathcal{L}^{4}_{n,i}\) are given by:
\begin{equation}
    \begin{aligned}
        \mathcal{L}^{3}_{n,i} &= - \log (1 - p_{n,i}), \\
        \mathcal{L}^{4}_{n,i} &= -(1 - q_3) \log (1 - p_{n,i}) - q_3 \log (p_{n,i}).
    \end{aligned}
\end{equation}

% Here, \(q_3\) is the label smoothing coefficient. The corresponding weight functions are defined as follows:
% \begin{equation*}
% \begin{aligned}
%     v^3(p; \alpha) &= \exp\left(-\frac{(p - \mu)^2}{2\sigma^2}\right), \\
%     v^4(p; \alpha) &= \min\left( 
%         \max\left( 
%             1 - \exp\left(-\frac{(p - \mu)^2}{2\sigma^2}\right), \lambda_{1} 
%         \right), 
%         \lambda_{2}
%     \right).
% \end{aligned}
% \end{equation*}

% \noindent where \(\lambda_1\) and \(\lambda_2\) are hyperparameters used to narrow the range of \(v^4\).

To provide a clearer understanding of our loss function's design, we outline how it handles various cases based on label status as follows.

\textbf{Confirmed Positives (\(\hat{y}_{n,i} = 1\)):}  The loss term \(\mathcal{L}^1_{n,i} \) is retained to ensure that the model assigns high probability to the true positive classes. 

\textbf{Undefined Pseudo-labels (\(\hat{y}_{n,i} = 0 \land l_{n,i} = 0\)):}  
The term \(\mathcal{L}^2_{n,i}\) (inherited from GR Loss) is used when no additional information from pseudo labeling is provided, and the label is naively assumed to be negative due to the high negative rate.

\textbf{Negative Pseudo-labels (\(\hat{y}_{n,i} = 0 \land l_{n,i} = -1\)):}  
The loss term \(\mathcal{L}^3_{n,i}\), unlike \(\mathcal{L}^2_{n,i}\) which includes the control of \(\hat{k}(p_{n,i})\), directly penalizes the model for assigning a high probability to classes that are considered negative by the pseudo-labels. Moreover, by employing the same weight function 
\(v^3(p;\alpha) = \exp\left(-\frac{(p-\mu)^2}{2\sigma^2}\right),\)
%\[
%v^3(p;\alpha) = \exp\left(-\frac{(p-\mu)^2}{2\sigma^2}\right),
%\]
which is also used in \(\mathcal{L}^2_{n,i}\) for undefined pseudo-labels, the method balances the contributions between positive and negative cases. 
    
\textbf{Positive Pseudo-labels (\(\hat{y}_{n,i} = 0 \land l_{n,i} = 1\)):}  
The loss term \(\mathcal{L}^4_{n,i}\) is a softened version of \(\mathcal{L}^1_{n,i}\) through the coefficient \(q_3\). This label smoothing helps to temper the model’s confidence, reducing overfitting to noisy pseudo-labels while still promoting the learning of positive associations when a pseudo-label is provided. The weight function 
\[
v^4(p;\alpha) = \min\Biggl( \max\Bigl( 1 - \exp\left(-\frac{(p-\mu)^2}{2\sigma^2}\right), \lambda_1 \Bigr), \lambda_2 \Biggr)
\]
constrains the influence of this term within a desirable range, controlled by \(\lambda_1\) and \(\lambda_2\).

For stable learning when introducing additional pseudo-labels, a regularization term \(\mathcal{R}\) is used to restrict the number of positive predictions with a coefficient \(\eta\), as in \cite{cole2021multi}:

\begin{equation}
   \mathcal{R} =\left(\frac{\hat{m}-m}{C}\right)^2,
\end{equation}
where \(m\) is the expected number of positive labels per image that can either be estimated from the available data or treated as a hyperparameter, such as: 

\begin{equation}
\label{eq:e_pos}
m \approx E_{pos} = \mathbf{E}_{(x, y) \sim p_{\text {data }}(x, y)} \sum_{i=1}^C \mathds{1}_{\left[y_i=1\right]},
\end{equation}
and \(\hat{m}\) represents the average sum of probabilities per image, computed as:
\begin{equation}
\label{eq:estimate_pos}
\hat{m}=\frac{1}{N}\sum_{n=1}^N \sum_{i=1}^C p_{n,i}.
\end{equation}

% \textbf{Motivation for the Regularization Term:}  
% The regularization term \(\mathcal{R}\) acts as a global calibration mechanism to ensure that the overall number of positive predictions remains consistent with the expected number derived from the data. This is particularly important in the single positive multi-label setting, where an overabundance of positive predictions due to noisy pseudo-labels can lead to degraded performance.

Our loss function \(\mathcal{L}^{GPR}\) is a generalization of \(\mathcal{L}^{GR}\) as shown in the following theorem.

\begin{theorem}[Our GPR Loss generalizes GR Loss]
\label{thm:gpr_to_gr}

Given a training dataset \(\{(x_n, \hat{y}_n, l_n)\}_{n=1}^{N}\) and a validation set \(\{(x^{val}_n, y^{val}_n)\}_{n=1}^{N'}\) as described in \cref{sec:problem_statement}, define:
\[
   \mathbf{C}(\mathcal{M}) = \exp \left( \sum_{n=1}^{N} \sum_{i=1}^{C} \mathds{1}_{[\hat{y}_{n,i} = 1]} \log P(l_{n,i} = 1 \mid x_n, \mathcal{M} ) \right),
\]
and:
\[
   m' = \frac{1}{N'} \sum_{n=1}^{N'} \sum_{i=1}^{C} \mathds{1}_{[y^{val}_{n,i} = 1]}.
\]
Then \(\mathcal{L}^{\text{GPR}}\) tends to \(\mathcal{L}^{\text{GR}}\) when \(\max \left( \mathbf{C}(\mathcal{M}), \left| m' - \hat{m} \right| \right)\) tends to zero.
Here, \(\hat{m}\) is defined in \cref{eq:estimate_pos}.
\end{theorem}

\begin{remark}
Intuitively, the above theorem says that our loss function \(\mathcal{L}^{\text{GPR}}\) will become the loss function \(\mathcal{L}^{\text{GR}}\) when \(\max \left( \mathbf{C}(\mathcal{M}), \left| m' - \hat{m} \right| \right)\) tends to zero. The quantity \( \mathbf{C}(\mathcal{M}) \) represents the confidence of method \( \mathcal{M} \) within the range \( (0, e] \) when generating pseudo-labels based on the given single positive label (\(\hat{y}_{n,i} = 1\)). Lower values of \( \mathbf{C}(\mathcal{M}) \), particularly those below \( \epsilon \), indicate a significant divergence between the pseudo-label probability distribution and the ground truth distribution, leading to \( l_{n,i} = 0 \) for all \( n, i \).
\end{remark}

\begin{remark}
When \(N'\) is sufficiently large, \(m' \to E_{pos}\), with \(E_{pos}\) defined in \cref{eq:e_pos}, then \(\hat{m} \to E_{pos}\). As a result, the regularization term in \cref{eq:gpr_loss} vanishes, i.e., \(\mathcal{R} \to 0\). This demonstrates that the GPR Loss naturally calibrates itself when the pseudo-label confidence and the expected positive label statistics align.
\end{remark}

The proof of this theorem can be found in the supplementary materials.

\section{DAMP}

In this section, we introduce a simple yet effective pseudo-labeling method that integrates with GPR Loss via a Dynamic Augmented Multi-focus Pseudo-labeling (DAMP) approach to form the framework AEVLP, as shown in \cref{fig:proposal}, for SPML. Further implementation details are provided in the supplementary materials.

\subsection{CLIP Inference and Strengthening with Noise}
\label{sec:clip_inference_noise}

As introduced in \cite{radford2021learning}, given an image input \( x \) and the \( i \)-th class from a set of \( C \) classes, the corresponding visual embedding and textual embedding are \( h = E_v(x) \in \mathbb{R}^K \) and \( t_i = E_t(\mathcal{P}_i) \in \mathbb{R}^K \), respectively. Here, \( E_v \) and \( E_t \) are the image and text encoders of the CLIP model with dimension \( K \), and \( \mathcal{P}_i \) is a predefined prompt for class \( i \), such as \textit{``a photo of a \{class\}''}. Let \( \hat{s}_i = \operatorname{Cos}(h,t_i) \), \( s_i = \operatorname{Softmax}\left(\frac{\hat{s}_i}{\tau}\right) \), and denote \( S = \{s_1, s_2, \ldots, s_C\} \) as the probability distribution for \( x \) across \( C \) classes.

Several works, including \cite{zhu2019freelb, kong2022robust, jain2023neftune}, have studied enhancing model performance during fine-tuning by adding controlled noise to model embeddings. Additionally, in \cite{ding2023exploring, chen2019multi}, label-to-label relationships are presented by GCN. Inspired by this, we propose adding controlled label-to-label correspondence noise to the text embeddings of the CLIP model. Concretely, we redefine the text embedding and update the GCN as: \( t_i = G(E_t(\mathcal{P}_i)) + E_t(\mathcal{P}_i) \) and \( H_{l+1} = \operatorname{LeakyReLU}(A^* H_l W_l) \) for \( l \in [0, L) \), with \( H_0 = \{E_t(\mathcal{P}_i) \mid 1 \leq i \leq C \} \). The graph \( G \) remains frozen during training, and the weights \( W_l \) are initialized from a uniform distribution. The adjacency matrix \( A^* \) is derived from the cosine similarity scores between the text embeddings of the classes produced by the CLIP text encoder.

\subsection{Dynamic Augmented Multi-focus Pseudo-labeling}
\label{sec:augmentor}

% By utilizing the method described in \cref{sec:zeroshot_clip} and following \cite{abdelfattah2023cdul}, we introduce a method that can generate more robust pseudo-labels in a multi-label setting through the use of multiple augmented views and single positive labels.
% In conjunction with the original image, we perform zero-shot classification using these patches and, finally, obtain probabilities from both global and local views, which are then aggregated to extract the final pseudo labels. Inspirited of this approach, we propose a novel CLIP-based pseudo-labeling from multiple augmented views and single positive labels. 

\paragraph{Augmentor.} %Let \(I\) be a given image of size \(H \times W\). We first divide the image \(I\) into a grid of size \(g \times g\). Each rectangle in the grid, which has nominal dimensions of \(\frac{H}{g} \times \frac{W}{g}\), is then expanded by a random ratio \(r\), forming a patch. This creates an overlap between adjacent patches. We then process the image and its patches using a transformation pipeline \(\operatorname{T}(\cdot)\), which includes standard preprocessing and weakly data augmentation techniques to generate various views for CLIP as follows: \(x^{global} = \operatorname{T}(I)\), \( x^{local}_{z} = \operatorname{T}(P_z)\). Let \(R = g^2\) be the total number of patches, and \(\{P_z\}_{z=1}^R\) be the patches.

Let \(I\) be an image. It is partitioned into overlapping patches \(\{P_z\}_{z=1}^R\) by dividing it into a grid and slightly enlarging each patch by a random ratio to ensure overlap. We then process the image and its patches using a transformation pipeline \(\operatorname{T}(\cdot)\), which includes standard preprocessing and weak data augmentation techniques, to generate various views for CLIP: the global view \(x^{global} = \operatorname{T}(I)\) and the local views \(x^{local}_z = \operatorname{T}(P_z)\). Following \cref{sec:clip_inference_noise}, these views yield the probability distributions \(S^{global} = \{s^{global}_1, s^{global}_2, \ldots, s^{global}_C\}\) and \(S^{local}_z = \{s^{local}_{z,1}, s^{local}_{z,2}, \ldots, s^{local}_{z,C}\}\).

\paragraph{Local threshold based on single positives.} Let \(\hat{c}\) be the given single positive label, according to the SPML setting in \cref{sec:problem_statement}, for the image \(I\). The local threshold \(\zeta^{local}\), which defines the patches to be trusted, is adjusted based on \( \zeta^{local} = \min(s_{\hat{c}}^{global}, \nu)\), where \(\nu\) is the general local threshold, set as a hyperparameter. In some cases, if \(\nu\) is set too high to recognize hard positives, we should consider the scores above \(s_{\hat{c}}^{global}\), as these can be meaningful, since \(\hat{c}\) is one of the true labels of the global view.

\paragraph{Aggregator.} From the local distributions \(\{S_z^{\text{local}}\}_{z=1}^R\), we aggregate a unified local distribution \(S^{\text{agg}}\) following \cite{abdelfattah2023cdul}. For each class \(c\), we compute \(\omega_c = \max_{z=1,\ldots,R} s_{z,c}^{\text{local}}\) and \(\psi_c = \min_{z=1,\ldots,R} s_{z,c}^{\text{local}}\), and define the aggregation score as \(s_c^{\text{agg}} = \mathds{1}_{[\omega_c \geq \zeta^{\text{local}}]}\omega_c + \mathds{1}_{[\omega_c < \zeta^{\text{local}}]}\psi_c\); this yields the soft aggregation vector \(S^{\text{agg}} = \{s_1^{\text{agg}}, s_2^{\text{agg}}, \ldots, s_C^{\text{agg}}\}\) for each input image.

\paragraph{Positive pseudo-labels.} To extract reliable positive pseudo-labels, we integrate both global and aggregated local similarities into \(S^{final} = \frac{1}{2} \left(S^{global} + S^{agg}\right)\). Let \(Q^{\prime} = \{l_1^{\prime}, l_2^{\prime}, \cdots,l_C^{\prime}\}\) be the pseudo labels of the image \(I\). We convert the soft similarity scores \(S^{final}\) into hard pseudo-labels as follows:
\begin{equation*}
l_c^{\prime} = \begin{cases}
1, & \quad s_c^{final} \in \operatorname{TopK}(S^{final}, k)  \And s_c^{final} \geq \zeta^{global} \\
0, & \quad \text{otherwise},
\end{cases}
\end{equation*}

\noindent where \(\zeta^{global}\) is the global threshold for high-confidence positive pseudo-labels, set as a hyperparameter, and \(k\) limits the number of positive pseudo-labels.

\paragraph{Negative pseudo-labels.} To identify potential negative pseudo-labels we compute average similarity scores as:

\begin{equation*}
    S^{avg} = \frac{1}{2} \left( S^{global} + \frac{1}{R}\sum_{z=1}^{R} S_{z}^{local} \right).
\end{equation*}

\noindent We use \(S^{\text{avg}}\) to refine \(Q^{\prime}\) by assigning negative pseudo-labels, producing the final pseudo-labels \(Q = \{l_1, l_2, \cdots, l_C\}\). A class \(c\) is designated as a negative pseudo-label if its score \(s_c^{\text{avg}}\) falls within the lowest \(\Delta_{\text{neg}} \%\) of values in \(S^{\text{avg}}\). Assuming a potential negative pseudo-label has low scores in both image \(I\) and every patch \(P_z\) according to the VLM, we define the assignment as:
\begin{equation*}
    l_c = \begin{cases} 
-1, & \quad s_c^{\text{avg}} \leq \theta_{\Delta_{\text{neg}}}(S^{\text{avg}}) \\ 
l_c', & \quad \text{otherwise} 
\end{cases}
\end{equation*}

\noindent where \(\theta_{\Delta_{\text{neg}}}(S^{\text{avg}})\) denotes the \(\Delta_{\text{neg}}\)-th-percentile of \(S^{\text{avg}}\), serving as the threshold to identify the lowest \(\Delta_{\text{neg}}\%\) of values in \(S^{\text{avg}}\) as negative pseudo-labels.

\section{Experiments}

\begin{table*}[htpb]
\centering
\setlength{\tabcolsep}{12pt}
\caption[Experimental results of AEVLP]{Experimental results on various benchmarks with SPML setting according to \cite{cole2021multi}. Our results are averaged over three runs as suggested in \cite{kim2023bridging}.}
\begin{tabular}{l|llll}
\hline
Method                       & VOC            & COCO           & NUS            & CUB            \\ \hline
Full-label (BCE)                  & 89.42          & 76.78          & 52.08          & 30.90          \\ \hline
Assume Negative (CVPR'21) \cite{cole2021multi}               & 85.89          & 64.92          & 42.27          & 18.31          \\
ROLE (CVPR'21) \cite{cole2021multi}              & 87.77          & 67.04          & 41.63          & 13.66          \\
EM (ECCV'22)     \cite{zhou2022acknowledging}            & 89.09          & 70.70          & 47.15          & 20.85          \\
EM + APL (ECCV'22)     \cite{zhou2022acknowledging}      & 89.19          & 70.87          & 47.59          & 21.84          \\
BoostLU + LL-R (CVPR'23)  \cite{kim2023bridging}   & 89.29          & 72.89          & \underline{49.59}    & 19.8           \\
SMILE (NeurIPS'22)    \cite{xu2022one}       & 86.31          & 63.33          & 43.61          & 18.61          \\
MIME (ICML'23)   \cite{liu2023revisiting}            & 89.20          & 72.92          & 48.74          & 21.89          \\
GR Loss (IJCAI'24) \cite{chen2024boosting} & \underline{89.83}    & \underline{73.17}    & 49.08          & 21.64          \\ 
VLPL (CVPRW'24)   \cite{xing2024vision}           & 89.10          & 71.45          & 49.55          & \underline{24.02}    \\ \hline
AEVLP (Ours)                 & \textbf{90.46} & \textbf{73.54} & \textbf{50.70} & \textbf{24.89} \\ \hline
\end{tabular}
\label{tab:result}
\end{table*}

\begin{table}[htpb]
\caption[]{Comparison of mAP scores when applying GPR Loss to different pseudo-labeling strategies.}
\label{tab:gpr_loss}
\resizebox{\columnwidth}{!}{
\begin{tabular}{l|cccc|c}
\hline
Method                     & VOC   & COCO  & NUSWIDE & CUB   & Average                               \\ \hline
BCE + Random & 86.84 & 67.46 & 41.23   & 10.69 & 51.56                                 \\
GPR + Random & 87.33 & 70.00 & 43.60   & 14.98 &  53.98 (+ 2.42)\\ \hline
VLPL                       & 89.1  & 71.45 & 49.55   & 24.02 & 58.53                                 \\
GPR + VLPL   & 89.52 & 72.83 & 49.93   & 24.22 & 59.13 (+ 0.6)                         \\ \hline
LL-Ct                      & 89.00 & 70.50 & 48.00   & 20.40 & 56.98                                 \\
GPR + LL-Ct  & 89.5  & 71.66 & 47.72   & 20.41 & 57.32 (+ 0.34)                        \\ \hline
BCE + DAMP   & 88.72  & 71.89 & 48.60   & 24.01 & 58.31 \\
GPR + DAMP   & 90.46  & 73.54 & 50.7   & 24.89 & 59.90 (+ 1.59) \\\hline
\end{tabular}
}
% \vspace{-0.5cm}
\end{table}

\subsection{Setup}
\label{sec:setup}
\paragraph{Datasets.} In this study, our method is evaluated through environmental experiments similar to those in \cite{cole2021multi, kim2022large, kim2023bridging, chen2024boosting, xing2024vision} across four standard benchmark datasets: PASCAL VOC 2012 (VOC) \cite{pascalvoc2012}, MS-COCO 2014 (COCO) \cite{coco2014}, NUS-WIDE (NUS) \cite{nus2009}, and CUB-200-2011 (CUB) \cite{cub2011}. From these fully labeled multi-label datasets, we simulate single positive training data by randomly retaining one positive label per training example, using the same seed as in \cite{cole2021multi}. Twenty percent of the training set for each dataset is set aside for validation purposes. Both the validation and test sets are fully labeled.
\paragraph{Implementation details.} For a fair comparison, we follow the standard SPML implementation described in \cite{cole2021multi, kim2023bridging, chen2024boosting, xing2024vision}, using the ResNet-50 architecture \cite{he2016deep}, pre-trained on ImageNet \cite{deng2009imagenet}. Each image is resized to \(448 \times 448\) and augmented with random horizontal flipping. For the VLPL method in \cite{xing2024vision}, we use CLIP ViT-L/16, while our approach uses CLIP ViT-B/16, both adopted from \cite{radford2021learning} to balance pseudo-labeling quality and training time. We train the models for 10 epochs on CUB and NUS, and 8 epochs on COCO and VOC, using the Adam optimizer \cite{kingma2014adam}. The batch size is set to 8 for CUB and VOC, and 16 for NUS and COCO. A learning rate of \(1 \times 10^{-5}\) is used across all experiments. We evaluate performance using the mean Average Precision (mAP) metric, conducting the final evaluation on the test set with the model that achieves the highest performance on the withheld validation set, in line with previous methods \cite{cole2021multi, kim2022large, kim2023bridging, chen2024boosting}.

\paragraph{Baselines.}
For a comprehensive understanding, we compare our proposed method, AEVLP, against several state-of-the-art methods within the SPML setting. Specifically, we evaluated against the following methods: Assume Negative \cite{cole2021multi}, ROLE \cite{cole2021multi}, EM \cite{zhou2022acknowledging}, EM + APL \cite{zhou2022acknowledging}, BoostLU + LL-R \cite{kim2023bridging}, SMILE \cite{xu2022one}, MINE \cite{liu2023revisiting}, GR Loss \cite{chen2024boosting}, VLPL \cite{xing2024vision}. All methods above, including our AEVLP, were evaluated on the same benchmark datasets: VOC, COCO, NUS, and CUB for a fair comparison. The results were averaged over three runs, as recommended in previous studies \cite{kim2023bridging}. 

% Note that we do not compare with the SCPNet method in \cite{ding2023exploring}. Despite the fact that both SCPNet and AEVLP methods are powered by Vision-Language models, SCPNet fine-tunes the CLIP model with a ResNet-50 backbone, initialized from the original OpenAI weights in \cite{radford2021learning}, rather than using the ResNet-50 backbone trained on ImageNet \cite{deng2009imagenet}. Additionally, SCPNet requires both image and text encoders during inference. 

\begin{figure*}[]
     \centering
     \begin{subfigure}[b]{0.33\textwidth}
         \centering
         \includegraphics[width=\textwidth]{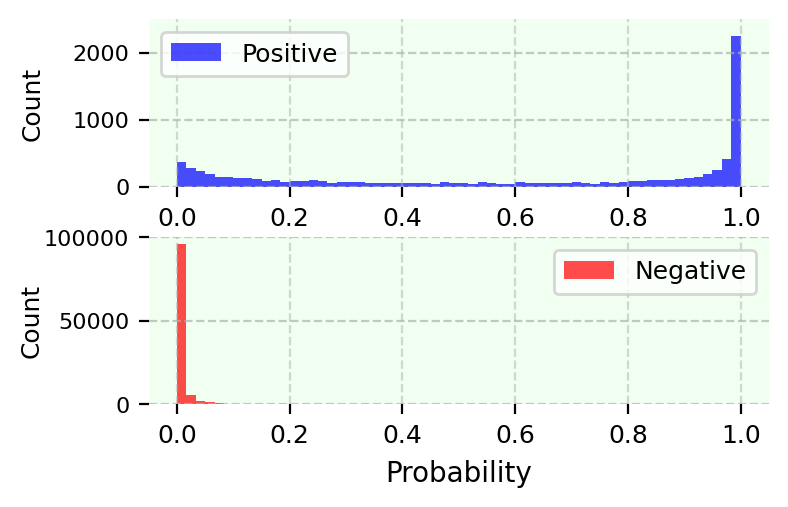}
         \caption{Assume Negative \cite{cole2021multi}}
         \label{fig:distribution_an}
     \end{subfigure}
     \hfill
     \begin{subfigure}[b]{0.33\textwidth}
         \centering
         \includegraphics[width=\textwidth]{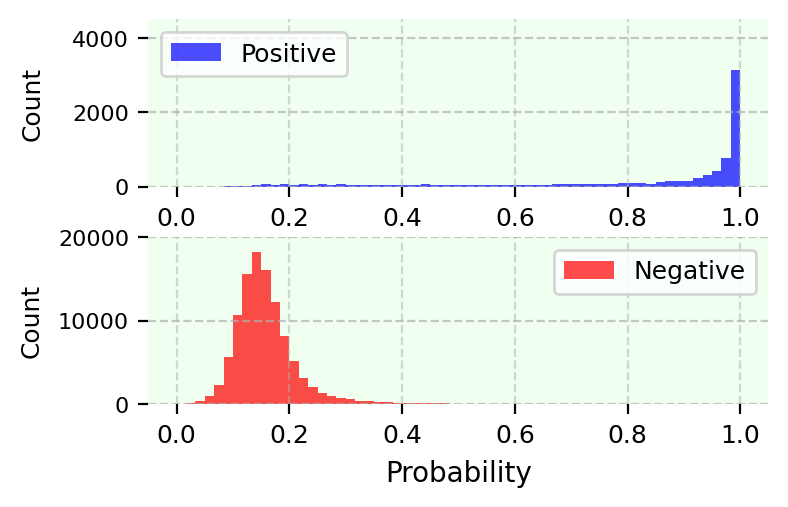}
         \caption{GR Loss \cite{chen2024boosting}}
         \label{fig:distribution_gr}
     \end{subfigure}
     \hfill
     \begin{subfigure}[b]{0.33\textwidth}
         \centering
         \includegraphics[width=\textwidth]{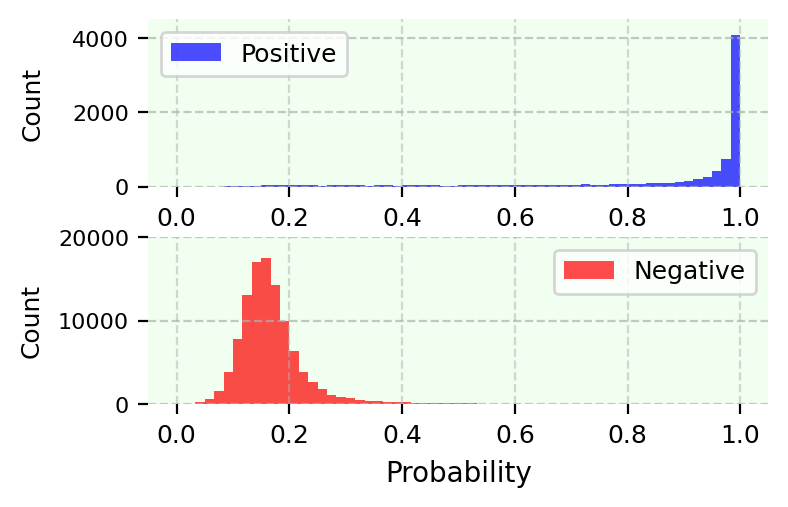}
         \caption{AEVLP}
         \label{fig:distribution_gpr}
     \end{subfigure}
        \caption{Distribution of output probabilities for the positive and negative classes of the VOC test set, as predicted by the ResNet-50 backbone classifier trained in the SPML setting.}
        \label{fig:distribution}
\end{figure*}

\subsection{Results and Discussion}

 Table \ref{tab:result} shows experimental results on four benchmark datasets: VOC, COCO, NUS, and CUB, using the previously described settings. Our method, AEVLP, achieves state-of-the-art performance across all datasets, demonstrating both effectiveness and robustness. Notably, it surpasses fully labeled multi-label classification on the VOC dataset. %Table \ref{tab:result} shows that our method consistently outperforms prior approaches across all benchmarks.
% Our conjecture is that this approach offers two advantages. Firstly, presenting different focuses of the same image in addition to the entire image to the VLM reduces its chance of missing certain classes present in the image. Secondly, repeatedly applying VLM to the same image but with  different randomized focuses throughout the training process allows a class being missed from the previous VLM runs to have a chance to be uncovered in the next runs. This has an error-averaging effect on the pseudo-label generation.
Although AEVLP can be seen as a fusion of GR Loss and VLPL, it not only combines their strengths but also surpasses both methods. VLPL, in particular, shows limitations in 3 out of 4 benchmark datasets compared with traditional state-of-the-art SPML methods. In contrast, AEVLP demonstrates superior performance, underscoring the effectiveness of introducing different focuses of the same input image to a VLM for pseudo-label generation. Notably, VLPL uses the CLIP model with a ViT Large backbone (428M parameters) \cite{dosovitskiy2021an}, with $14\times14$ patches and a 336-pixel resolution, whereas AEVLP achieves better results while using the CLIP model with a smaller ViT Base backbone (149.62M parameters) with $16\times16$ patches and a $224\times224$ input resolution (ViT-B/16). As can be seen from Table \ref{tab:result}, traditional SPML methods \cite{cole2021multi, xu2022one, liu2023revisiting, chen2024boosting} based on pseudo-labeling approach struggle with datasets containing a large number of labels, such as CUB (312 labels). AEVLP, in contrast, shows a relative improvement of more than 15\% over GR Loss method \cite{chen2024boosting} and 3.62\% over the next best-performing method, VLPL \cite{xing2024vision}, extending GR Loss's capability to work with pseudo-labels. Our proposed GPR Loss, described in \cref{sec:gprloss}), can cope with pseudo-labels generated by various mechanisms, as long as they provide an acceptable level of confidence.
% for example, the CLIP model.

\begin{table}[]
\caption[]{DAMP performance on several benchmark datasets.}
\label{tab:damp}
\resizebox{\columnwidth}{!}{
\begin{tabular}{l|cccc}
\hline
Metrics               & VOC   & COCO  & NUS   & CUB   \\ \hline
Average Precision     & 65.13 & 84.79 & 37.09 & 19.07 \\
Accumulated Precision & 60.33 & 81.59 & 34.49 & 18.29 \\ \hline
Average Recall        & 48.29 & 24.68 & 8.84  & 11.37 \\ 
Accumulated Recall    & 51.1  & 26.84 & 10.06 & 13.88 \\ \hline
\end{tabular}
}
% \vspace{-0.5cm}
\end{table}

\begin{table*}[htpb]
\caption{Ablation study on the main components of the AEVLP framework: noise addition \( G(\cdot) \), augmentation \( \operatorname{T}(\cdot) \), overlapping ratio \( r \), loss re-weighting for positive pseudo-labels \( v^4(p; \alpha) \), regularization term \( \mathcal{R} \), and negative pseudo-labels loss \( \mathcal{L}^3_{n,i} \). Note that \ding{55} for \( v^4(p; \alpha) \) indicates it is set to a constant, and for \( r \), it indicates \( r = 0 \).}
\label{tab:aevlp_components}

\setlength{\tabcolsep}{12pt}
\centering
\begin{tabular}{cccccc|cccc}
\hline
\multicolumn{6}{c|}{AEVLP}                                                           & \multicolumn{4}{c}{Datasets}                                                               \\ \hline
\multicolumn{3}{c|}{DAMP}                        & \multicolumn{3}{c|}{GPR Loss}      & \multirow{2}{*}{VOC} & \multirow{2}{*}{COCO} & \multirow{2}{*}{NUS} & \multirow{2}{*}{CUB} \\ \cline{1-6}
\(G(\cdot) \) & \(\operatorname{T}(\cdot)\) & \multicolumn{1}{c|}{\(r\)} & \(v^4(p; \alpha)\) & \(\mathcal{R}\) & \(\mathcal{L}^3_{n,i}\) &                      &                       &                      &                      \\ \hline
\ding{55}     & \ding{51}    & \multicolumn{1}{c|}{\ding{55} }           & \ding{55}      & \ding{55}              & \ding{55}        & 90.28                & 73.07                 & 50.00                & 24.14                \\
\ding{51}      & \ding{51}     & \multicolumn{1}{c|}{\ding{55}}           & \ding{55}      & \ding{55}              & \ding{55}        & 90.21                & 73.37                 & 50.32                & 24.27                \\
\ding{51}      & \ding{55}    & \multicolumn{1}{c|}{\ding{51} }           & \ding{55}      & \ding{55}              & \ding{55}        & 90.13                & 73.32                 & 50.46                & 24.15                \\
\ding{51}      & \ding{51}     & \multicolumn{1}{c|}{\ding{51} }           & \ding{55}      & \ding{55}              & \ding{55}        & 90.22                & 73.14                 & 50.57                & 24.24                \\
\ding{51}      & \ding{51}     & \multicolumn{1}{c|}{\ding{51} }           & \ding{51}       & \ding{55}              & \ding{55}        & 90.31                & 73.40                 & 50.37                & 24.73                \\
\ding{51}      & \ding{51}     & \multicolumn{1}{c|}{\ding{51} }           & \ding{51}       & \ding{51}              & \ding{55}        & 90.38                & 73.51                 & 50.64                & 24.83                \\
\ding{51}      & \ding{51}    & \multicolumn{1}{c|}{\ding{51} }           & \ding{51}       & \ding{51}               & \ding{51}         & \textbf{90.46}                & \textbf{73.54}                 & \textbf{50.70}                & \textbf{24.89}                \\ \hline
\end{tabular}
% \vspace{-0.5cm}
\end{table*}

\subsection{Analysis}
\paragraph{GPR Loss robustness.} GPR Loss is designed to work with various pseudo-labeling strategies, not only by effectively utilizing pseudo-labels but also by demonstrating robustness to noise. To validate this, we apply GPR Loss to pseudo-labels generated by different methods, including DAMP (AEVLP), Random Pseudo-labels, VLPL \cite{xing2024vision}, and LL-Ct \cite{kim2022large}. Random Pseudo-labels are obtained by randomly selecting unknown labels (assumed negative) as pseudo-positive labels, based on the average number per image in the dataset. VLPL provides pseudo-labels generated from the CLIP model, while LL-Ct converts assumed negative labels, particularly those with high loss values, into pseudo-positive labels. For the baseline of DAMP and Random Pseudo-labels, we use BCE for training; for the other pseudo-labeling methods, we replace their original loss function with GPR Loss. As shown in \cref{tab:gpr_loss}, incorporating GPR Loss consistently enhances mAP scores across all datasets for each pseudo-labeling strategy. This highlights GPR Loss's effectiveness in mitigating the inherent noise in pseudo-labels, regardless of the generation method.

%As shown in \cref{tab:gpr_loss}, for each pseudo-labeling strategy, incorporating GPR Loss consistently improves the mAP scores across all datasets. This demonstrates that GPR Loss effectively mitigates the noise inherent in pseudo-labels, regardless of how they are generated.

\paragraph{DAMP performance.}

We apply pseudo-labeling to the training data of the VOC, COCO, NUS, and CUB datasets using the DAMP technique, evaluating over the same number of epochs used for training the main results in \cref{tab:result}. Since pseudo-labels of a given image may vary across epochs, we report the results as average precision and recall across all epochs to assess the supervision quality that DAMP provides for training the backbone. For strict evaluation and to address label imbalance, we focus only on the missing positive labels in the training data according to the SPML setting, ignoring predictions for the confirmed single positive and true negative labels. To validate coverage of missing positive labels, we accumulate predictions across epochs: each new positive pseudo-label prediction is recorded, and precision and recall are computed on these accumulated positive labels. As illustrated in \cref{tab:damp}, DAMP, a strategy of randomization, enhances positive label coverage, with accumulated recall exceeding average recall, while maintaining a stable quality of pseudo-labels. Notably, precision across epochs varies only slightly from the average precision (see supplementary materials). In contrast, using fixed pseudo-labels results in lower coverage, and a straightforward accumulation approach introduces more noise.

\paragraph{Probability distribution analysis.} \Cref{fig:distribution} plots the distribution of the classifier output probabilities on the VOC test set from three different methods: Assume Negative, GR Loss, and AEVLP (ours). The ResNet-50 backbone is used in all three methods. Compared to existing method in \cref{tab:result}, the Assume Negative method represents a naive approach, while GR Loss shows robustness across all datasets. As shown in \cref{fig:distribution_an}, the Assume Negative method performs the worst: although it maintains high confidence for negative labels (close to 0), it includes numerous low-confidence predictions for positive labels, which can lead to significant errors. For both GR Loss in \cref{fig:distribution_gr} and AEVLP in \cref{fig:distribution_gpr}, the probability distributions of negative labels have a similar shape, concentrated around 0.2. However, AEVLP outperforms GR Loss, with its distribution of positive labels shifted closer to 1, even when learning from pseudo-labels that may introduce noise.
 
\begin{figure}[]
    \centering        
    \includegraphics[width=\columnwidth]{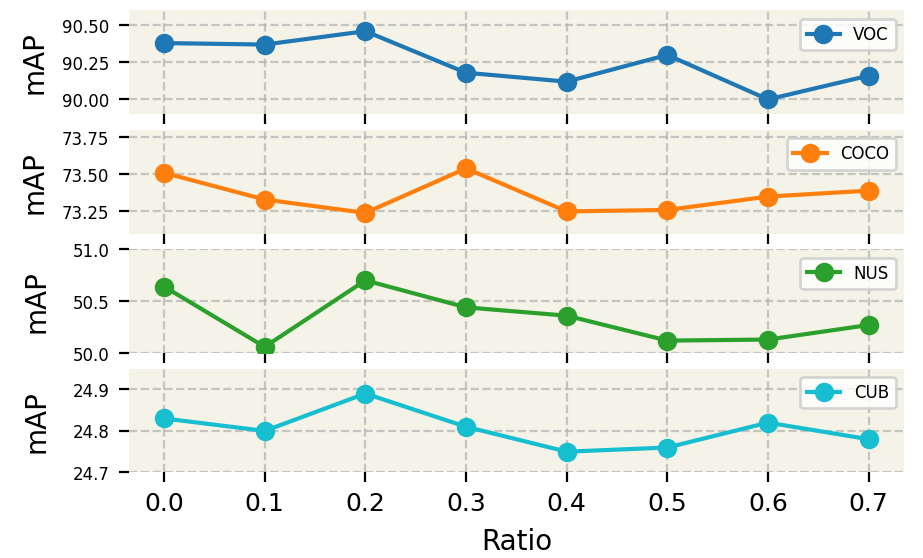}      \captionsetup{justification=centering, width=\columnwidth}
    %\caption[]{The performance of the model in learning from varying ratios of negative pseudo-labels across different datasets using the AEVLP method.}
    \caption[]{The performance of the model in learning from varying ratios of negative pseudo-labels across different datasets using the AEVLP method.}
    \label{fig:mining_negatives}
    % \vspace{-0.5cm}
\end{figure}

\paragraph{Mining negative pseudo-labels.} \cref{fig:mining_negatives} shows the impact of varying the negative pseudo-label ratio \(\Delta_{neg}\%\) on model performance. At approximately \(20\%\) to \(30\%\)  negative pseudo-labels, the model achieves its best mAP performance across all datasets. This suggests that introducing a moderate proportion of negative pseudo-labels helps improve diversity in the learning process without significantly increasing noise. As the negative pseudo-label ratio increases further, the mAP slightly decreases but remains relatively stable. This is likely due to the influence of GPR Loss, which helps mitigate the negative impact of potentially inaccurate pseudo-labels in situations where there is an imbalance between positive and negative labels.

\subsection{Ablation Study} 

\paragraph{Components of AEVLP.} In this section, we analyze the impact of each component in the AEVLP framework, as shown in \cref{tab:aevlp_components}. The ablation results in \cref{tab:aevlp_components} shows that each component contributes to the overall performance improvement. Some observations are relevant here to describe the relationships between closely linked components that complement each other. The combination of noise addition \( G(\cdot) \) with either augmentation \( \operatorname{T}(\cdot) \) or overlapping \( r \) provides a noticeable boost, as these pairs introduce valuable pseudo-labels that may be difficult to capture with the standard input alone. The positive pseudo-label re-weighting \( v^4(p; \alpha) \) and regularization term \( \mathcal{R} \) help stabilize learning from uncertain pseudo-labels. Additionally, the negative pseudo-label loss \( \mathcal{L}^3_{n,i} \) encourages a stronger focus on potential negative labels, thereby enhancing discrimination.

\begin{table}[]
\centering
\caption{mAP scores for various grid sizes \(g\) on VOC and CUB datasets.}
\begin{tabular}{l|ccccc}
\hline
Grid size & 2     & 3     & 4              & 5              & 6     \\ \hline
VOC       & 90.13 & 90.08 & \textbf{90.46} & 90.21          & 90.21 \\ \hline
COCO       & 73.07 & 73.13 & \textbf{73.54} & 73.48          & 73.32 \\ \hline
NUS       & 50.25 & 50.19 & \textbf{50.7} & 50.35          & 50.15 \\ \hline
CUB       & 24.37 & 24.38 & 24.48          & \textbf{24.89} & 24.67 \\ \hline
\end{tabular}
\label{tab:grid_size_ablation}
% \vspace{-0.1cm}
\end{table}

% \paragraph{Impact of the grid size.} We evaluate the impact of different grid sizes \(g\) (ranging from 2 to 6) on model performance using two datasets: VOC (20 labels) and CUB (312 labels), representing contrasting label quantities. The results, shown in \cref{tab:grid_size_ablation}, highlight the highest mAP scores in bold. For the VOC dataset, a grid size of 4 yields the best performance, while for CUB, grid size 5 performs best. These findings suggest that the optimal grid size is proportional to the number of labels in the dataset, though it is also influenced by training resources. As the grid size increases, the number of labels per patch tends to decrease, which may improve label dominance, but a balance is required. The patches must remain large enough to capture even small objects; beyond a certain grid size, patches may become too small, reducing the method’s effectiveness. Additionally, larger grid sizes increase computational costs, as the number of patches grows quadratically. These results highlight the need to balance between label reduction, object capture, and computational efficiency when selecting the grid size.

\paragraph{Impact of the grid size.} 
We evaluate grid sizes \(g\) (from 2 to 6) on four datasets: VOC (20 labels), COCO (80 labels), NUS (81 labels), and CUB (312 labels). The results (see \cref{tab:grid_size_ablation}) show that a grid size of 4 yields the best performance for VOC, COCO, and NUS, suggesting that a moderate grid size effectively balances spatial detail and computational efficiency. In contrast, CUB achieves its highest mAP with a grid size of 5, indicating that datasets with finer-grained labels benefit from slightly larger grid sizes to capture richer object details. While increasing grid size reduces the number of labels per patch, enhancing label separability, excessively small patches may fail to capture entire object instances and incur higher computational costs. These findings underscore the need to balance label reduction, object capture, and efficiency when selecting the optimal grid size.

\paragraph{Backbone architecture. } 

\begin{table}[]
\caption[]{AEVLP performance on various backbone architectures. Note that 1K and 22K refer to pretrained models on ImageNet-1K and ImageNet-22K, respectively.}
\label{tab:arch}
\resizebox{\columnwidth}{!}{
\begin{tabular}{l|c|c|c|c}
\hline
Backbone & ResNet-50 & ConvNeXt-L-1K & ConvNeXt-L-22K & ViT-L/14       \\ \hline
VOC     & 90.46     & 92.92         & \textbf{93.11} & 93.10          \\ \hline
COCO     & 73.54     & 76.14         & \textbf{82.69} & 81.39          \\ \hline
NUS      & 50.7      & 50.9          & 54.74          & \textbf{54.98} \\ \hline
CUB      & 24.89      & \textbf{25.26}          & 24.87          & 23.36 \\ \hline
\end{tabular}
}
% \vspace{-0.5cm}
\end{table}
We evaluate several backbone architectures, as shown in \cref{tab:arch}. Specifically, we replace the baseline backbone with ConvNeXt-L \cite{liu2022convnet}, using two different pretrained versions (ImageNet-1K and ImageNet-22K), as well as ViT-L/14 \cite{radford2021learning}. The results demonstrate a notable improvement in performance with these architectures. This analysis confirms AEVLP's flexibility, showing it can effectively incorporate advanced models to achieve substantial performance gains.

% đo năng lực pseudo-labeling 
% for 1 --> N epoch: 
%     for 1 --> B minibatch: size of  (batchsize)
%         P dự đoán bộ nhãn giả y_pseudo_batch
%     ==> y_pseudo_epoch so khớp y_true_train không xét nhãn dương given
%     ==> avg precision, recall ==> pseudo-labeling stable as avg precision không thay đổi nhiều qua từng epoch
% general_precision = avg precision / n, genral_recall
% OR ==> accept tất cả y_pseudo_batch ==> bộ accumulated pseudo-labels 
% ==> accumulated recall > general_recall ==> tiếp cận nhiều nhãn dương hơn nhưng mà giữ được precision ổn định acc

% 100 missing 

% chạy 1 lần làm nhãn luôn thì 27 

% chạy randomly mỗi lần thì cuối training mô hình đã tiếp cận được 30  

% chạy N lần ramdomly, voting mechanism => heuristic 
% lần 1: chó, mèo 
% lần 2: chó, mèo, gà
% lần 3: chó 
% ==> chó, 
 
\section{Conclusion}

%This work addresses the challenging problem of SPML by introducing the AEVLP framework. The framework leverages CLIP's exceptional zero-shot classification capabilities and the robustness of the GPR Loss function. This approach reduces reliance on fully annotated datasets while overcoming the limitations of traditional multi-label learning methods.
% Dynamic Augmented Multi-focus Pseudo-labeling
In this paper, we propose AEVLP, a novel framework for SPML. First, our approach introduces a new GPR loss function that effectively complements the pseudo-labels. We then present a dynamic augmented multi-focus pseudo-labeling strategy designed to overcome the limitations of fixed pseudo-labeling approaches, where the pseudo-labels for each image remain constant throughout training. Experimental results demonstrate that AEVLP achieves state-of-the-art performance on four benchmark datasets, reducing reliance on fully annotated data while addressing the shortcomings of traditional multi-label learning methods.

{
    \small
    \bibliographystyle{ieeenat_fullname}
    \bibliography{main}
}
% \newpage
% \input{ICCV2025-Author-Kit-Feb/sec/X_suppl}

% {
%     \small
%     \bibliographystyle{ieeenat_fullname}
%     \bibliography{main}
% }
\end{document}